# Automatic Summarization of Doctor-Patient Encounter Dialogues Using Large Language Model through Prompt Tuning

**Mengxian Lyu, MS[1], Cheng Peng, PhD[1], Xiaohan Li, MS[1], Patrick Balian, DBA[1,2], Jiang Bian, PhD[1,2], Yonghui Wu, PhD[1,2*]**

**[1]Department of Health Outcomes and Biomedical Informatics, College of Medicine, University of Florida, Gainesville, FL, USA; [2]Cancer Informatics Shared Resource, University of Florida Health Cancer Center.**

## Abstract

*Automatic text summarization (ATS) is an emerging technology to assist clinicians in providing continuous and coordinated care. This study presents an approach to summarize doctor-patient dialogues using generative large language models (LLMs). We developed prompt-tuning algorithms to instruct generative LLMs to summarize clinical text. We examined the prompt-tuning strategies, the size of soft prompts, and the few-short learning ability of GatorTronGPT, a generative clinical LLM developed using 277 billion clinical and general English words with up to 20 billion parameters. We compared GatorTronGPT with a previous solution based on fine-tuning of a widely used T5 model, using a clinical benchmark dataset MTS-DIALOG. The experimental results show that the GatorTronGPT-20B model achieved the best performance on all evaluation metrics. The proposed solution has a low computing cost as the LLM parameters are not updated during prompt-tuning. This study demonstrates the efficiency of generative clinical LLMs for clinical ATS through prompt tuning.*

## Introduction

Clinicians constantly collect, interpret, and summarize critical patient information in electronic health records (EHRs) to facilitate the practice of medicine to provide continuous and coordinated care[1]. Appropriately summarizing their dialogue with the patient and documenting the patient's clinical care information is a critical skill for clinicians to communicate with patients (e.g., discharge summaries), handoff work at the change of shift, and work with colleagues with different expertise and across different clinical departments to handle complex cases (e.g., patients with multiple illnesses). However, the growing burden of clinical documentation has been a significant challenge in healthcare, contributing to physician burnout, reducing the efficiency of care delivery, and potentially compromising patient safety[2–4]. Automatic text summarization (ATS) is a promising technique to assist clinicians in summarizing patient information from extensive data from a variety of sources to alleviate the documentation workload and improve continuous and coordinated care[5].

ATS is a natural language processing (NLP) task to produce a text, i.e., a summary, from one or more original texts; the summary is expected to capture the key information from and much shorter than the original text. Typically, ATS can be categorized into extractive summarization and abstractive summarization. Extractive summarization involves selecting certain portions of the original text to create a summary, essentially extracting the most important information without changing the original text[6]. In contrast, abstractive summarization generates a new piece of text that reflects the essence of the original information concisely and coherently, which requires a deeper understanding of source materials[7]. Based on the number of documents in the original text, ATS can also be classified as single-document summarization - where the ATS generates a summary from a single document, or multi-document summarization - where there are multiple documents in the original text[8].

Various solutions including statistical-based, knowledge-based, topic-based, and machine learning-based models have been extensively explored for the ATS[8]. Traditional machine learning-based solutions approach ATS as a sentence-level binary classification task, aiming to classify each sentence of a document as 'positive' (i.e., to include in the summary) or 'negative' (i.e., to exclude from the summary). Recent advancements in NLP have demonstrated the potential of pre-trained Large Language Models (LLMs) across all NLP tasks[9]. LLMs are machine learning models based on the deep neural network architecture known as transformers[10]. Through pre-training on massive amounts of text, LLMs demonstrated better transfer learning, few-shot learning, and zero-shot learning abilities, enabling them to tackle nearly all NLP subtasks[11,12]. Based on whether the LLM used the encoder component or the decoder component of a transformer model, LLMs can be categorized into encoder-only, decoder-only, and encoder-decoder models. LLMs that contain a decoder component (i.e., the decoder-only and encoder-decoder LLMs) are considered generative LLMs as they can generate coherent and contextually relevant text such as emails, articles, and even computer codes. Generative LLMs have demonstrated strong performance in NLP tasks that require long context input, such as question answering (QA) and ATS. LLMs approach ATS as a text-to-text generation task that aims to generate a summary from the original text, which greatly improves abstractive summarization. Generative LLMs based on GPT[13]

*Corresponding author, Yonghui.Wu@ufl.edu

(Generative Pre-trained Transformer, a decoder-only model) and T5[14] (Text-to-Text Transfer Transformer, an encoder-decoder model), have been effectively applied in ATS.

Prompt-based learning is the key technology that utilizes a 'prompt'—additional instructional information added to the input data—to guide LLMs in generating text that follows these instructions. There are two types of prompts: (1) "hard prompts" (or discrete prompts), which are manually crafted text containing specific instructions or information about the NLP tasks. The use of hard prompts to instruct LLMs to generate answers of interest is referred to as hard prompting; and (2) "soft prompts" (or continuous prompts), which are a continuous vector (of virtual tokens) added into the input[15]. The use of soft prompts in LLMs is referred to as soft prompting or prompt tuning. Hard prompting requires prompt engineering, a time-consuming process that entails manually creating and experimenting with many variations of discrete prompts and requires an intricate understanding of the model's underlying mechanisms[15]. Hard prompting also requires updating all billions of model parameters, leading to high computing costs. Conversely, in soft prompting, the model parameters are frozen (i.e., not updated during fine-tuning) and only the trainable soft prompts are updated during fine-tuning to learn task-specific prompts, thereby unloading researchers from the labor-intensive task of prompt engineering[16]. Soft prompting also has a low computing cost, as only small, trainable prompts are updated during fine-tuning while the LLM parameters are kept frozen[17]. Previously, we have developed clinical LLMs including GatorTron[18] and GatorTronGPT[19] , and have explored soft prompting for multiple clinical NLP tasks and demonstrated good performance in clinical concept extraction, clinical relation extraction, clinical abbreviation disambiguation, natural language inference, medication attribute filling, clinical concept normalization, and progress note understanding[20–22].

This study seeks to develop a cost-efficient ATS method for summarizing doctor-patient dialogue into clinical notes using the GatorTronGPT LLM through prompt tuning. We examined various prompt tuning strategies involving LSTM (long short-term memory) and multilayer perceptron (MLP) models, the length of soft prompts, and the few-short learning ability of generative LLMs for ATS. We also compared our GatorTronGPT-based system with a widely used generative LLM, T5. The experimental results, using a widely recognized benchmark dataset for clinical ATS, MTS-DIALOG[23], demonstrated the effectiveness of LLMs in generating succinct and clinically relevant summaries of clinical narratives. Our approach differs from previous studies that used LLMs through hard prompting with fine-tuning, where there is a need to update a massive amount of model parameters, resulting in a very high computational cost. The proposed method offers a cost-efficient solution for instructing generative LLMs to summarize clinical text.

## Methods

### Dataset

This study used the MTS-DIALOG dataset[23], a widely recognized benchmark dataset for clinical ATS, which consists of 1,701 doctor-patient encounter dialogues and their corresponding clinical note sections, covering a broad spectrum of medical specialties and note types. The dataset was compiled from de-identified real-world clinical notes summarized by eight domain experts with medical backgrounds. The dataset is organized into a training set of 1201 samples, a validation of 100 samples, and a test set of 400 samples. There are six categories of clinical note types: General Medicine (1,035 notes, 60.8%), Neurology (296 notes, 17.4%), Orthopedic (208 notes, 12.2%), Subjective, Objective, Assessment and Plan (SOAP) (79 notes, 4.6%), Dermatology (56 notes, 3.3%), and Allergy/Immunology (27 notes, 1.6%). **Table 1** shows the statistical summary of the MTS-DIALOG dataset, while **Table 2** shows an example of a doctor-patient dialogue and the corresponding note from the dataset.

**Table 1**. Statistics of the MTS-DIALOG Dataset.

| | **Datasets** | **Sample Number** | **Dialogue Average Word Count** | **Summary Average Word Count** |
|---|---|---|---|---|
| MTS-DIALOG | Training | 1201 | 104.12 | 39.95 |
| | Validation | 100 | 100.35 | 38.45 |
| | Test | 400 | 102.95 | 39.05 |

**Table 2.** Example of doctor-patient dialogue and the corresponding clinical note section from MTS-DIALOG Dataset

| **Doctor-Patient Dialogue** | **Clinical Note Section** |
| --- | --- |
| **Doctor:** What brings you back into the clinic today, miss?<br>**Patient:** I came in for a refill of my blood pressure medicine.<br>**Doctor:** It looks like Doctor [**{NAME}**] followed up with you last time regarding your hypertension, osteoarthritis, osteoporosis, hypothyroidism, allergic rhinitis and kidney stones. Have you noticed any changes or do you have any concerns regarding these issues?<br>**Patient:** No.<br>**Doctor:** Have you had any fever or chills, cough, congestion, nausea, vomiting, chest pain, chest pressure?<br>**Patient:** No.<br>**Doctor:** Great. Also, for our records, how old are you and what race do you identify yourself as?<br>**Patient:** I am seventy six years old and identify as a white female. | The patient is a 76-year-old white female who presents to the clinic today originally for hypertension and a med check. She has a history of hypertension, osteoarthritis, osteoporosis, hypothyroidism, allergic rhinitis and kidney stones. Since her last visit she has been followed by [**{NAME}**]. Those issues are stable. She has had no fever or chills, cough, congestion, nausea, vomiting, chest pain, chest pressure. |

### Doctor-patient Dialogue Summarization as text-to-text Generation

We approached doctor-patient dialogue summarization as a text-to-text generation task, using the entire dialogue text as input. To instruct LLMs to generate summaries, we attached soft prompts to the input, as trainable variables, which were randomly initiated at the beginning and optimized during prompt tuning. Specifically, we formulated clinical ATS as a prompt-based text-to-text generation task: given an input dialogue $D = d_1, \ldots, d_k$, where $d_k$ denotes the $k_{th}$ token within the input, the goal is to train soft prompts that can instruct LLMs generating an output $R = r_1 \ldots, r_n$, which encapsulates the important information but be shorter than D, as an abstractive ATS task.

### Large Language Models

This study explored LLMs based on two mainstream architectures including the GPT and T5.

- **GatorTronGPT:** GatorTronGPT[19] is a generative clinical LLM based on the GPT-3 architecture. We pretrained GatorTronGPT with 277 billion words of text, incorporating 82 billion words of de-identified clinical texts from the University of Florida (UF) Health system and 195 billion words of general English from the Pile[25] dataset. GatorTronGPT is developed to enhance clinical NLP and healthcare text generation[19]. Our previous study has demonstrated its ability to generate text that pass the clinical experts' Turing tests for linguistic readability and clinical relevance, making it a valuable tool for clinical ATS. We used two different versions of GatorTronGPT: GatorTronGPT-5B, a GPT model trained with 5 billion parameters; and GatorTronGPT-20B, a GPT model trained using 20 billion parameters.
- **T5**: Text-to-Text Transfer Transformer, known as T5[11], is an encoder-decoder Transformer model pre-trained on the "Colossal Clean Crawled Corpus". T5 models adopt a unified text-to-text framework, where all NLP tasks are considered as converting one form of text into another. T5 models are pretrained across a variety of tasks including machine translation, question answering, text classification, and summarization. In this study, we utilized the Huggingface fine-tuning pipeline[a] to finetune a T5-Large variant using the MTS-DIALOG dataset. All parameters of T5-Large were updated during the fine-tuning process.

[a] https://github.com/huggingface/transformers/tree/main/examples/pytorch/summarization

**Prompt Tuning Using Soft Prompts**

To develop a computational cost-efficient method, we adopted prompt tuning to add "soft prompts" as trainable variables in the input to guide the GatorTronGPT in generating summaries of interests. During prompt tuning, we kept our model weights frozen and only updated the parameters of the soft prompts. We used a soft prompt template, which incorporates a placeholder '<|VIRTUAL_PROMPT|>' and includes trainable embeddings within the prompt structure. The template is structured as follows: '<|VIRTUAL_PROMPT|> Input: {input}\n Output:{output}', where the placeholder is designed to be replaced with a sequence of virtual tokens during the training phase. These virtual tokens are associated with embeddings dynamically updated through Multi-Layer Perceptron (MLP) and Long Short-Term Memory (LSTM) networks. There are two sections in the prompt template: the 'Input:' marker, indicating the start of the input dialogue, and the 'Output:' marker, used to guide the model to generate a response. During both the training and inference stages, the '{input}' placeholder is replaced with actual input data, while the '{output}' segment is utilized to frame the model's generated response, which, however, is omitted during the inference phase. We used a cross-entropy loss function to calculate the loss based on the discrepancy between the model-generated summary and the gold-standard summary, which was used to adjust the soft prompts through standard backpropagation.

**Figure 1.** Proposed Framework using Soft Prompt Tuning for Doctor–Patient Dialogue Summarization. This figure illustrates the architecture used for prompt tuning in doctor-patient dialogue summarization. The framework introduces virtual tokens that serve as soft prompts. These tokens are processed by a Prompt Encoder, which updates the virtual tokens via backpropagation during training. The model operates by receiving the embedded prompts and producing the final Output Summary. This architecture ensures the soft prompts are updated during training to optimize the model's performance.

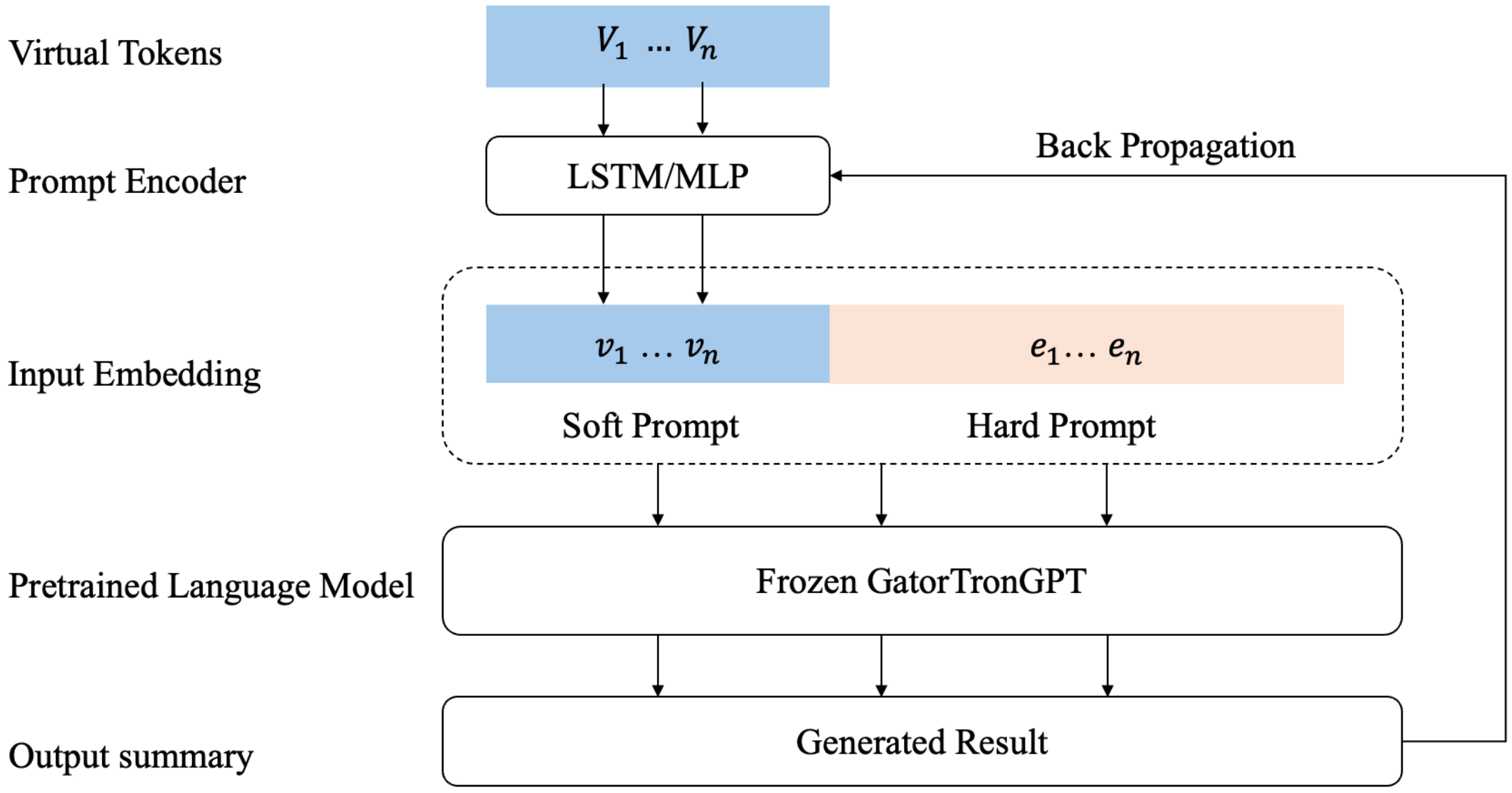

Frozen: The weight of GatorTronGPT is not changed during prompt tuning.

**Experiments and Evaluation**

We followed the standard prompt tuning procedure to tune the soft prompts using the training set while keeping GatorTronGPT models frozen, identifying the best parameters using the validation set, and evaluating the performance using the test set. We conducted a grid search to identify the optimal combination of the number of encoder layers, the hidden sizes, and the strategies to initialize the soft prompts (i.e., MLP and LSTM). We also experimented with different lengths of soft prompt tokens to identify the best lengths. We examined both the 5-billion-parameter (GatorTronGPT-5B) and 20-billion-parameter (GatorTronGPT-20B) GatorTronGPT models to examine how model sizes affect summarization quality. We also investigated the few-shot learning ability of the GatorTronGPT-20B model to assess its efficiency in low-data resource settings. A stratified sampling method was used to construct few-shot learning datasets, specifically, we selected the samples stratified by 20 different section headers used in the dataset until the corresponding sections had been used up.
We used standard evaluation metrics for ATS, including Rouge[25] (Recall-Oriented Understudy for Gisting Evaluation), which emphasize recall by measuring the overlap of N-grams, longest common subsequences, or skip-bigrams between the generated and reference texts, thus capturing different aspects of text overlap; BLEU[26] (Bilingual Evaluation Understudy), focuses on N-gram generation accuracy, assessing the precision of the generated text; and BERTScore[27](Microsoft/deberta-xlarge-mnli), which employs embeddings from the BERT model between the generated and reference texts, effectively capturing deeper semantic relationships and contextual coherence. A T5-Large variant finetuned on the same datasets is used as the baseline. All parameters of the T5-Large were updated during the fine-tuning.

**Experiment Settings**
We developed the prompt tuning methods to instruct GatorTronGPT models using the Nvidia NeMo[28] package based on Python. We experimented with two strategies to initialize the soft prompts including LSTM (10 layers, 1024 hidden size) and MLP. A fused Adam optimizer was used for prompt tuning. We used grid search to identify the best layer sizes, hidden sizes, and learning rates. The best performance is achieved using an LSTM strategy with a layer size of 10 and a hidden size of 1024, a learning rate of 0.0001, and a CosineAnnealing scheduler to adjust the learning rate, a warm-up of 50 steps to ramp up gradually. For the T5-Large[14] model, we used the default Google/T5-Large model host by the official HuggingFace model repository and fine-tuned the model using the same training set and validation set. We used a maximum of 20 epochs for both the prompt tuning of GatorTronGPT models and the fine-tuning of T5-Large models and selected the best model checkpoint based on the validation performance on the validation set. We also optimized the parameters controlling the text-to-text generation and the best performance was achieved using top-k filtering of 1 and nucleus sampling (top-p) of 0.9, a temperature of 0.1. The maximum output token size was set to 64 based on the dataset statistic. All training experiments were conducted using 8 Nvidia A100-80G GPUs.

**Results**
**Table 3** compares prompt-tuning of GatorTronGPT models with different sizes of virtual tokens from 32 to 512 using evaluation metrics including Rouge-1, Rouge 2, Rouge-L, BERTScore (calculated using Microsoft/deberta-xlarge-mnli), and BLUE. The best size for virtual tokens was determined using the mean average overall evaluation scores. Both GatorTronGPT-5B and GatorTronGPT-20B achieved the best results using a virtual token size of 128. GatorTronGPT-20B outperformed GatorTronGPT-5B on all evaluation metrics. The GatorTronGPT-20B is more sensitive to the length of virtual tokens than GatorTronGPT-5B. As shown in Table 3, GatorTronGPT-20B had a larger performance gain than GatorTronGPT-5B when increasing the virtual token sizes.

**Table 3**. Comparison of GatorTronGPT models using different virtual token sizes.

| **Model** | **Virtual Token** | **Performance** | | | | | |
|---|---|---|---|---|---|---|---|
| | | **Rouge-1** | **Rouge-2** | **Rouge-L** | **BLEU** | **BERTScore** | **Overall** |
| GatorTronGPT-5B | 32 | 0.3348 | 0.1302 | 0.3195 | 0.3240 | 0.6902 | 0.3597 |
| | 64 | 0.3274 | 0.1260 | 0.3143 | 0.3245 | 0.6972 | 0.3579 |
| | 128 | 0.3359 | **0.1342** | **0.3221** | **0.3383** | **0.6993** | **0.3660** |
| | 256 | 0.3339 | 0.1296 | 0.3205 | 0.3245 | 0.6929 | 0.3603 |
| | 512 | **0.3385** | 0.1312 | 0.3215 | 0.3342 | 0.6933 | 0.3637 |
| GatorTronGPT-20B | 32 | 0.3546 | 0.1370 | 0.3361 | 0.3506 | 0.7102 | 0.3777 |
| | 64 | 0.3547 | 0.1408 | 0.3384 | 0.3526 | 0.7180 | 0.3809 |
| | 128 | **0.3628** | **0.1549** | **0.3472** | **0.3665** | **0.7309** | **0.3925** |
| | 256 | 0.3525 | 0.1515 | 0.3377 | 0.3427 | 0.7181 | 0.3805 |
| | 512 | 0.3495 | 0.1339 | 0.3323 | 0.3452 | 0.7165 | 0.3755 |

**Table 4**. Comparison of GatorTronGPT models with T5 for automatic summarization of clinical text.

| **Model** | **Initiate Method** | **Trainable Parameters** | **Training Duration** | **Performance** | | | | |
|---|---|---|---|---|---|---|---|---|
| | | | | **Rouge-1** | **Rouge-2** | **Rouge-L** | **BLEU** | **BERT Score** |
| T5-Large | Fine-tuning | 770M | 9h34m | 0.3425 | 0.1465 | 0.3368 | 0.3425 | 0.6765 |
| GatorTronGPT-5B | Prompt tuning | 70M | 2h10m | 0.3359 | 0.1342 | 0.3221 | 0.3383 | 0.6993 |
| GatorTronGPT-20B | Prompt tuning | 302M | 4h23m | **0.3628** | **0.1549** | **0.3472** | **0.3665** | **0.7309** |

**Table 4** compares the best GatorTronGPT-5B and GatorTronGPT-20B models with the widely used T5 model on trainable parameters and the evaluation performance calculated using the test set. GatorTronGPT-20B model achieved the best scores on all evaluation metrics, outperforming GatorTronGPT-5B and T5-Large. GatorTronGPT-5B achieved a better BERTScore (0.6993) than T5 (0.6765) but slightly lower scores for other evaluation metrics. In terms of the trainable parameters, GatorTronGPT-5B has only 70 million parameters to update in the prompt tuning, which is the lowest computational cost among all models. GatorTronGPT-20B used 302 million parameters in the prompt tuning, which is less than half of the T5 model with 770 million parameters. Therefore, prompt-tuning of GatorTronGPT-5B and GatorTronGPT-20B took 2 hours 10 minutes and 4 hours 23 minutes, respectively, which were remarkably computationally efficient than fine-tuning of the T5-Large model, which took 9 hours and 34 minutes.

**Table 5.** Examples of summaries generated by GatorTronGPT-20B and T5-Large

| | **Gold-Standard Summary** | **GatorTronGPT-20B Summary** | **T5 Summary** |
|---|---|---|---|
| 1 | The patient is a 79-year-old African-American female with a height of 5 foot 3 inches and weight of 197 pounds. She was diagnosed with type 2 diabetes in 1983. She is not allergic to any medicines. | The patient is a 79-year-old African-American female who is 5'3" tall and weighs 197 pounds. She has a history of type 2 diabetes mellitus since 1983. She has no known drug allergies. | This is a 79-year-old African American female who is self-reported 5\'3" tall and 197 pounds. She has a history of type 2 diabetes since 1983. She has no known allergies. |
| 2 | Bactrim, which causes nausea and vomiting, and adhesive tape. | Bactrim, which causes nausea and vomiting, adhesive tape. | Bactrim and adhesive tape. |
| 3 | MSK: Negative myalgia, negative joint pain, negative stiffness, negative weakness, negative back pain. | MUSCULOSKELETAL: No myalgias, arthralgias, stiffness, or weakness. No back pain. | Muscle pain, no joint pain, no stiffness or weakness. No back pain. |
| 4 | Father died of a thoracic aortic aneurysm, age 71. Mother died of stroke, age 81 | His father died of thoracic aortic aneurysm at age 71. His mother died of stroke at age 81. | Father died of thoracic aortic aneurysm at age 71. Mother died of stroke at age 81. |
| 5 | This is a 66-year-old patient who came to the emergency room because she was feeling dizzy and was found to be tachycardic and hypertensive. | This 66-year-old female presents to emergency room complaining of dizziness. She was found to have a high blood pressure and a tachycardia. | Hypertension and tachycardia. |

**Table 5** shows 5 examples of summaries randomly selected from GatorTronGPT-20B and T5, compared with the corresponding ground truth summaries. The summaries generated by GatorTronGPT-20B captured more critical information in the ground truth across various scenarios, including patient demographics to specific clinical conditions. The summaries generated by GatorTronGPT are more precise than the summaries generated by T5. For example, GatorTronGPT captured “no known drug allergies”, whereas T5 only captured “no known allergies” in example 1. GatorTronGPT also captured more critical conditions such as “nausea and vomiting” in example 2 and “positive for stoke” in example 5.

**Table 6.** Results of few-shot learning performance of GatorTronGPT-20B.

| Model | Sample | Rouge-1 | Rouge-2 | Rouge-L | BERTScore | BLEU |
|---|---|---|---|---|---|---|
| GatorTronGPT-20B | 5 | 0.1390 | 0.0374 | 0.1308 | 0.3525 | 0.0646 |
| | 10 | 0.1823 | 0.0576 | 0.1720 | 0.4579 | 0.1102 |
| | 20 | 0.1897 | 0.0620 | 0.1806 | 0.6390 | 0.1893 |
| | 40 | 0.2245 | 0.0823 | 0.2138 | 0.6363 | 0.1486 |
| | 60 | 0.2731 | 0.1012 | 0.2620 | 0.6807 | 0.2501 |
| | 100 | 0.3007 | 0.1163 | 0.2899 | 0.6899 | 0.2811 |
| | 200 | 0.3164 | 0.1241 | 0.2980 | 0.6977 | 0.2959 |
| | Full dataset:1201 | 0.3628 | 0.1549 | 0.3472 | 0.7309 | 0.3665 |

**Table 6** evaluates the few-shot learning ability of the GatorTronGPT-20B model. We compared the performance of GatorTronGPT-20B prompt-tuned using different numbers of samples from 5 to 200 and compared with the best performance derived using the full training set of 1,201 examples. Not surprisingly, all evaluation scores improved when increasing the number of samples for prompt tuning. GatorTronGPT achieved decent performance using 200 samples compared with the best performances derived using the entire training set.

**Discussion and Conclusions**

Clinical ATS is a promising technology to assist clinicians in clinical documentation. In this study, we developed NLP methods to instruct LLMs in summarizing doctor-patient dialogues into clinical notes through prompt tuning. This method is cost-efficient and relieves end-users from the labor-intensive prompt engineering process of using LLMs. We systematically examined strategies to initialize virtual tokens for prompt tuning, the size of virtual tokens, the computing cost, and the few-shot learning ability of LLMs through prompt tuning. The evaluation results using a benchmark clinical ATS dataset show that the proposed prompt tuning method based on the GatorTronGPT-20B model achieved the best evaluation scores on all metrics, demonstrating the effectiveness of generative LLMs for clinical ATS.

The proposed method has a low computing cost. Previous studies widely used fine-tuning to train LLMs for ATS, where all the parameters of the LLM will be updated, which has a high computing cost. As LLMs are getting larger, fine-tuning LLMs is time-consuming and has high computing costs. Early studies exploring prompt tuning with smaller LLMs have reported that prompt tuning has a remarkable performance gap to be comparable with fine-tuning. This study explored a much larger generative LLM, GatorTronGPT, with up to 20 billion parameters. The experimental results show that by updating only a small fraction of parameters (70 million) through prompt tuning, we can achieve performance close to the traditional fine-tuning approaches with 770 million parameters updated in the fine-tuning. By updating 302 million parameters, GatorTronGPT-20B achieved the best scores on all evaluation metrics outperforming the fine-tuning of the T5 model. Our results demonstrated that prompt tuning of generative LLMs can achieve comparable or even better performance than traditional fine-tuning for clinical ATS. However, prompt tuning requires large-size generative LLMs to achieve performance comparable to traditional fine-tuning.

We experimented with different sizes of virtual tokens in prompt tuning and tested the few-shot learning ability of GatorTronGPT. The experimental results show that the size of virtual tokens did impact model performance. Larger generative LLMs are more sensitive to the sizes than smaller LLMs. Both two GatorTronGPT models achieved the best performance using a size of 128, which is consistent with our previous research exploring GatorTronGPT for clinical concept extraction. The few-shot learning results show that by using 200 samples, GatorTronGPT-20B can achieve decent performance on BERTScore, but there are still large gaps for other evaluation metrics. This also indicates that the few-shot learning ability of GatorTronGPT can be further improved.

This study has limitations. Automatic evaluation methods, such as Rouge and BERTScore, may not be good enough to assess the quality of summarization, as they measure the N-gram overlap and semantic similarity. These metrics provide objective scores to evaluate LLMs for ATS, but they may not be fully aligned with human judgment. The few-shot learning ability of GatorTronGPT can be further improved using reinforcement learning from human feedback (RLHF) or multi-task instruction tuning.

**Acknowledgment**

This study was partially supported by a Patient-Centered Outcomes Research Institute® (PCORI®) Award (ME-2018C3-14754), the PARADIGM program awarded by the Advanced Research Projects Agency for Health (ARPA-H), a grant from the National Cancer Institute, R01CA246418, grants from the National Institute on Aging, NIA R56AG069880, R01AG080624, R01AG083039, R01AG080991, R01AG084236, R01AG084178, R01AG076234, National Institute of Allergy and Infectious Diseases, NIAID R01AI172875, National Heart, Lung, and Blood Institute, R01HL169277, the Cancer Informatics Shared Resource supported by the UF Health Cancer Center and the UF Clinical and Translational Science Institute Biomedical Informatics Program. The content is solely the responsibility of the authors and does not necessarily represent the official views of the funding institutions.